\documentclass[runningheads]{llncs}

\usepackage[]{eccv}
\usepackage{tabularray}

\usepackage{eccvabbrv}

\usepackage{graphicx}
\usepackage{booktabs}

\usepackage[accsupp]{axessibility}  

\usepackage{graphicx}
\usepackage{subcaption}
\usepackage{colortbl}
\usepackage{pifont}
\usepackage{caption}

\usepackage[pagebackref,breaklinks,colorlinks]{hyperref}

\usepackage{orcidlink}

\begin{document}

\title{HGS-Mapping: Online Dense Mapping Using Hybrid Gaussian Representation in Urban Scenes} 

\titlerunning{Abbreviated paper title}

\author{Ke Wu\textsuperscript{\rm 1}, 
        Kaizhao Zhang\textsuperscript{\rm 2}, 
        Zhiwei Zhang\textsuperscript{\rm 1}, 
        Shanshuai Yuan\textsuperscript{\rm 1}, 
        Muer Tie\textsuperscript{\rm 1}, 
        Julong Wei\textsuperscript{\rm 1}, 
        Zijun Xu\textsuperscript{\rm 1}, 
        Jieru Zhao\textsuperscript{\rm 3}, 
        Zhongxue Gan\textsuperscript{\rm 1}, 
        Wenchao Ding\textsuperscript{\rm 1}}

\authorrunning{F.~Author et al.}

\institute{Fudan University \email{23110860017@m.fudan.edu.com}\\ \and
School of Future Technology, Harbin Institute of Technology\\  \and
Department of Computer Science and Engineering, Shanghai Jiao Tong University}

\maketitle

\begin{abstract}

Online dense mapping of urban scenes forms a fundamental cornerstone for scene understanding and navigation of autonomous vehicles. Recent advancements in mapping methods are mainly based on NeRF, whose rendering speed is too slow to meet online requirements. 3D Gaussian Splatting (3DGS), with its rendering speed hundreds of times faster than NeRF, holds greater potential in online dense mapping. However, integrating 3DGS into a street-view dense mapping framework still faces two challenges, including incomplete reconstruction due to the absence of geometric information beyond the LiDAR coverage area and extensive computation for reconstruction in large urban scenes. To this end, we propose HGS-Mapping, an online dense mapping framework in unbounded large-scale scenes. To attain complete construction, our framework introduces Hybrid Gaussian Representation, which models different parts of the entire scene using Gaussians with distinct properties. Furthermore, we employ a hybrid Gaussian initialization mechanism and an adaptive update method to achieve high-fidelity and rapid reconstruction. To the best of our knowledge, we are the first to integrate Gaussian representation into online dense mapping of urban scenes. Our approach achieves SOTA reconstruction accuracy while only employing 66\% number of Gaussians, leading to 20\% faster reconstruction speed.




\end{abstract}

\section{Introduction}
\label{sec:intro}

Online dense mapping is crucial for autonomous vehicles to understand complex environments and navigate effectively. The meaning of online dense mapping lies in continuously creating a detailed map of the vehicle's surroundings using sensor data streams (camera and LiDAR, etc.) as the vehicle moves. This detailed map encompasses almost all visible surfaces and objects, offering a comprehensive and high-fidelity representation of the environment.

\begin{figure}[t] 
    \centering 
    \includegraphics[width=\textwidth]{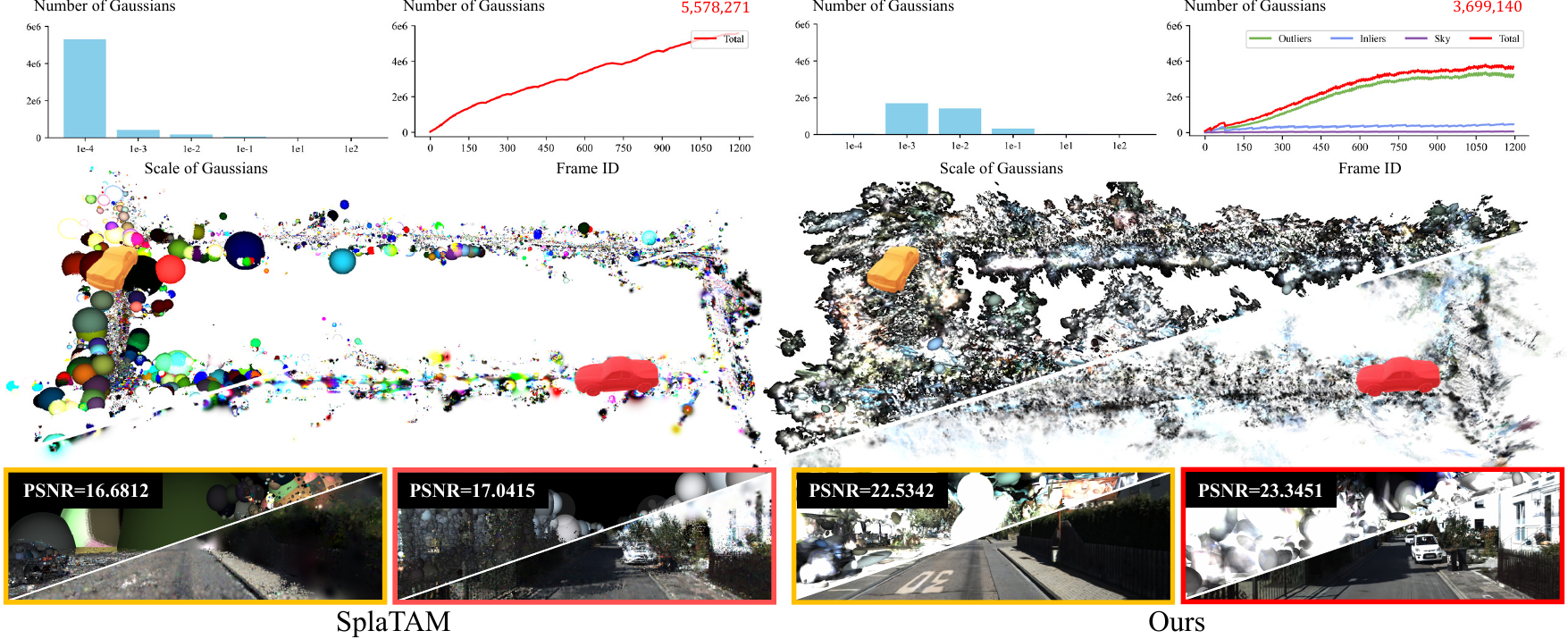} 
    \caption{\textbf{Reconstruction of large scale urban scenes.} Our method achieves high-quality rendering results using only two-thirds the number of Gaussians compared to the current SOTA online reconstruction method SplaTAM~\cite{splatam}.}
    \label{cover} 
    \vspace{-0.7cm}
\end{figure}

How to represent this dense map has been studied for decades. Conventional methods~\cite{vins,rtabmap,orbslam2,elasticfusion} directly fuse sensor data across space and time to construct a map, but this kind of map is sparse, unable to capture rich scene details, and contains holes. In recent years, NeRF-based methods have gained popularity due to their realistic novel view synthesis. 
NeRF~\cite{nerf} uses resource-intensive neural networks to represent the color and density of the scene and outputs the entire image in a volume rendering manner, which leads to several training hours. Subsequent methods~\cite{instantngp,niceslam,coslam} accelerate training and rendering speed by combing voxel features with light MLPs, introducing NeRF into online mapping. However, these methods are limited by the finite resolution of voxels, making it difficult to balance rendering quality and speed. Moreover, the inherent volume rendering process of NeRF requires sampling all the pixels, which is the fundamental reason for deficient training and rendering speed.

3D Gaussian Splatting (3DGS)~\cite{3DGS} has received much attention in recent months due to its high-fidelity rendering quality and rendering speed that is hundreds of times faster than NeRF, holding greater potential in online dense mapping. 3DGS utilizes Gaussians with learnable parameters to represent the entire scene, replacing NeRF's volume rendering with splatting-based rasterization~\cite{ewasplatting}. Despite the impressive rendering quality and speed of 3DGS, research on using 3DGS representation for online dense mapping is scarce and entirely limited to indoor scenarios~\cite{splatam,gaussian-slam,gaussian-splatting-slam,gs-slam}. This limitation arises because 3DGS requires pre-processing feature point clouds with Structure-from-Motion (SfM)~\cite{sfm} on a complete dataset for initializing Gaussians, which is not feasible for online dense mapping tasks. Existing methods in indoor scenarios typically use a depth stream to initialize Gaussians. In complex outdoor environments, depth cameras often struggle to function effectively, leading to challenges in obtaining dense depth information for urban scenes. The straightforward approach is to initialize Gaussians using LiDAR points. Nonetheless, introducing 3D Gaussian representation into online dense mapping in urban scenes still faces the following two gaps. Firstly, LiDAR points can only cover a limited range; for example, the upper parts of trees and buildings, and distant landscapes may lack LiDAR coverage. Consequently, this results in a lack of initial geometry information for these areas, making it impossible for existing Gaussian-based methods to fully reconstruct the scene. Secondly, large urban scenes require a significant number of Gaussians, leading to substantial memory overhead and computational complexity during the rasterization process.

To tackle these challenges outlined above, we present HGS-Mapping, an online dense mapping framework that fully leverages the potential of Gaussian representation, achieving accurate and efficient reconstruction for entire urban environments at the same time. Our approach introduces a novel representation called Hybrid Gaussian Representation, comprised of Sphere Gaussian, 3D Gaussian, and 2D Gaussian Plane components. This representation achieves high-fidelity reconstruction results for the areas lacking LiDAR points. 
Thus, the entire urban scene is fully represented in this manner. Furthermore, we implement an adaptive update method for Gaussians, which dynamically densifies Gaussians based on the reconstruction loss and prunes the Gaussians of low importance. This update method enhances rendering quality and accelerates rendering speed. As shown in \cref{cover}, we utilize merely 66\% of the number of Gaussians in SplaTAM~\cite{splatam} and optimize Gaussians to an appropriately non-minimal scale. With fewer Gaussians, we successfully convey richer scene information and render images of higher quality. \textit{To the best of our knowledge, we are the first to integrate Gaussian representation into online dense mapping for urban scenes.}
%

Our contributions can be summarized as follows:
\begin{itemize} 
    \item We are the first to introduce Gaussian representation to online dense mapping in urban scenes, and we innovatively propose a Hybrid Gaussian Representation, which facilitates reconstructing entire urban scenes. 
    \item We propose a hybrid Gaussian initialization mechanism and an adaptive update method, which result in superior rendering quality and faster reconstruction speeds.
    \item Comprehensive experiments demonstrate that HGS-Mapping surpasses previous methods in both rendering quality and speed while employing only two-thirds the number of Gaussians compared to the SOTA approach.
\end{itemize}

\section{Related Works}
\textbf{\textit{3D Gaussian Splatting.}} 3DGS~\cite{3DGS} employs 3D Gaussians to represent scenes and achieves fast differentiable rendering through splat-based rasterization. Numerous downstream tasks have emerged, such as driving scenes simulation~\cite{drivinggaussian,streetgaussian}, simultaneous localization and mapping~\cite{splatam,gs-slam,gaussian-slam,gaussian-splatting-slam}, mesh reconstruction \cite{neusg,sugar}, and dynamic scene modeling \cite{4dgaussiansplatting,dynamic3dgaussians,yang2023real}, etc. To reduce storage capacity, LightGaussian~\cite{lightgaussian} computes the global significance of Gaussians and prunes away those below a certain threshold. Gaussianpro~\cite{gaussianpro} introduces a progressive propagation strategy to guide Gaussian densification and tests on the Waymo dataset~\cite{waymo}.
However, current online mapping methods built on 3D Gaussians predominantly target indoor scene reconstruction, overlooking mechanisms suitable for outdoor environments. Additionally, most efforts in reconstructing scenes for autonomous driving rely on SfM~\cite{sfm} for Gaussian initialization, a process which is too slow to satisfy the real-time demands of online dense mapping.


\noindent\textbf{\textit{Urban Scale Mapping.}} With the success of NeRF~\cite{nerf} in 3D reconstruction, an increasing number of studies are focusing on reconstruction and simulation for autonomous driving in urban scenes~\cite{unisim,lightsim,streetsurf,emernerf,neural-lidar-fields,urban-radiance-fields,suds,neuras}. However, due to the vastness, unbounded nature, and difficult accessibility of depth information in urban environments, urban scale mapping presents significant challenges. For example, to address the sky regions in urban scenes where rays never intersect with any opaque surfaces,~\cite{urban-radiance-fields,streetsurf} propose an MLP to map the ray direction to the color of the sky. For dynamic objects within the urban scene, Unisim~\cite{unisim} adopts the NSG~\cite{nsg} approach of representing dynamic scenes as scene graphs for editing purposes. Streetsurf~\cite{streetsurf} employs the signed distance function (SDF) to portray urban scenes and draws inspiration from the methodology of NeRF++~\cite{nerf++} to decompose unbounded space using a hyper-cuboid model. ~\cite{streetgaussian,drivinggaussian} firstly employ 3D Gaussian splatting for the reconstruction of urban scenes. Specifically, DrivingGaussian~\cite{drivinggaussian} utilizes a dynamic Gaussian graph to describe the scene dynamically and incorporates LiDAR priors for an incremental sequential background representation. StreetGaussian~\cite{streetgaussian} employs a 4D spherical harmonics model to characterize dynamic objects and uses semantic logits for rendering semantic images. However, there hasn't been any work yet applying advanced Gaussian representation to online mapping in urban environments.

\section{Method}

\begin{figure}[t] 
    \centering 
    \includegraphics[width=\textwidth]{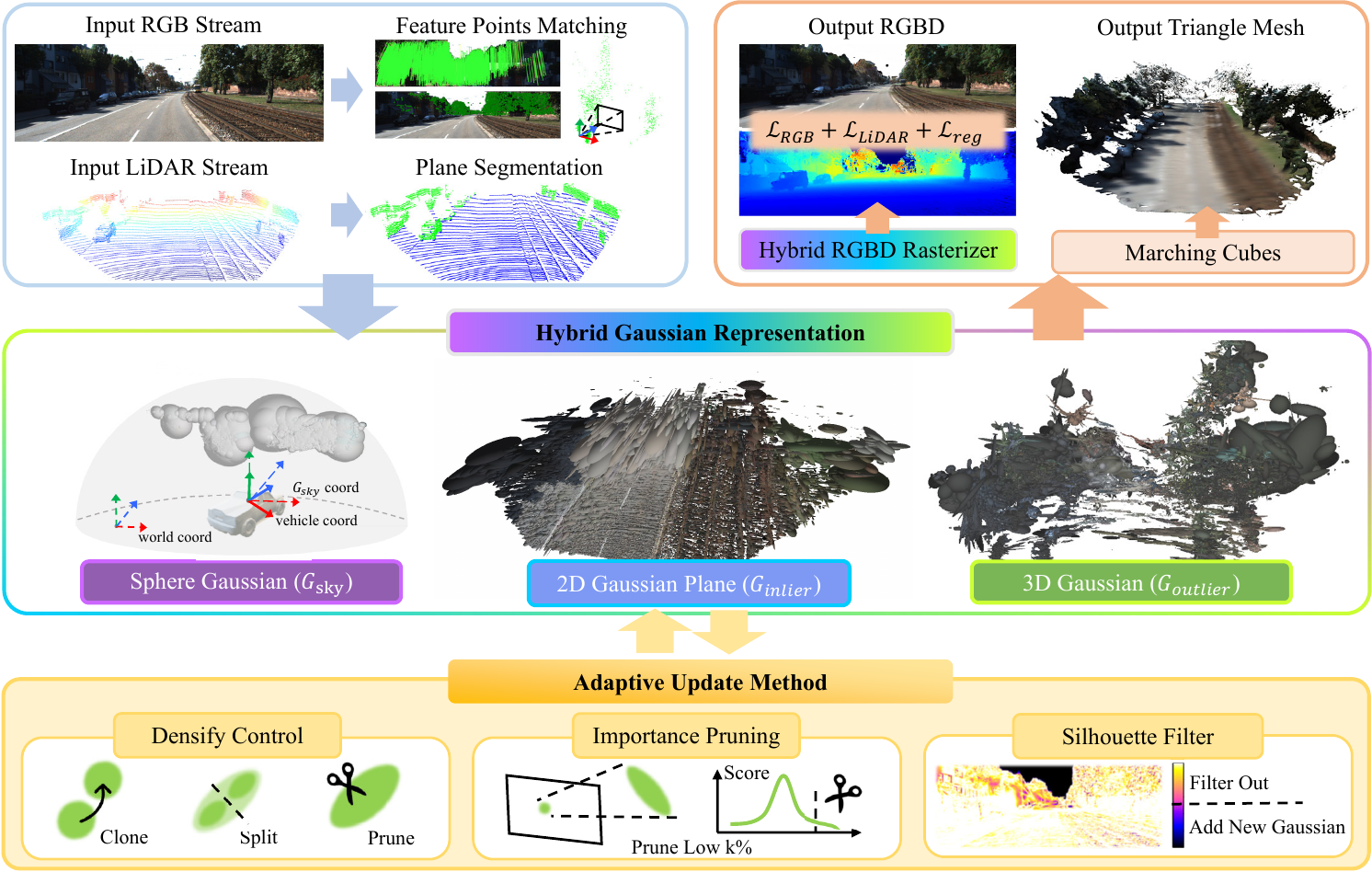} 
    \caption{\textbf{Overview of HGS-Mapping's pipeline.} Input RGB-LiDAR stream and pose sequence are incrementally integrated into our Hybrid Gaussian Representation. We update and adjust the Gaussians with our Adaptive Update Method (Densify Control, Importance Pruning, Silhouette Filter). Our method supports rendering RGB-Depth using Hybrid Gaussian Rasterizer and exporting meshes.} 
    \label{pipeline} 
    \vspace{-0.5cm}
\end{figure}

The overview of our pipeline is illustrated in \cref{pipeline}. The input of our system includes RGB-SparseD (LiDAR) stream $\{I_t, L_t\}$ along with a pose sequence ${T_t}$ from a moving vehicle in urban streets. To initialize Gaussians, we match the feature points from adjacent RGB frames with the corresponding LiDAR points. Based on the Hybrid Gaussian Representation (\cref{hybrid_gaussian_representation_text}), we model the world as inliers, outliers, and the sky separately. We design an efficient rasterizer tailored for such representation (\cref{rasterization_and_optimization_text}). Following this, we use an adaptive update method 
 (\cref{adaptive_update_method_text}) to incrementally maintain and optimize our Hybrid Gaussian Representation. Our framework supports rendering high-quality RGB and depth images and exporting meshes using marching cubes.

\subsection{Hybrid Gaussian Representation}\label{hybrid_gaussian_representation_text}
It is instinctive for human to divide the world into the sky, the ground, and ground objects due to their significant differences in geometry and photometry. Consequently, using the same representation to model these different parts is inefficient.  We propose Hybrid Gaussian Representation consisting of three types of Gaussians: $G_{sky}$, $G_{inlier}$ and $G_{outlier}$, which are used to model different parts of the urban scenario, namely the sky, the road surface, and roadside scenery respectively. Our Hybrid Gaussian Representation records the color and geometric information of the scene. The initialization method and the properties for each type of Gaussians are distinct.

\subsubsection{Gaussian Initialization.}\label{gaussian_init} Vanilla 3DGS initializes Gaussians with a sparse point cloud obtained by running Structure-from-Motion (SfM) on the complete dataset. However, this is infeasible for online mapping processes in which the vehicle is mapping while moving, and the very sparse viewpoints can also lead to incorrect geometric structures generated by SfM. 

To achieve better initialization, we use LiDAR points to initialize Gaussians. However, LiDAR's coverage is quite limited, unable to capture the upper parts of trees and buildings, as well as distant landscapes.

To this end, for areas within the LiDAR's coverage, we initialize Gaussians using LiDAR points. For the remaining areas, we implement a lightweight feature matching network~\cite{lightglue} to extract matching pixels from adjacent RGB frames and calculate optical flow value $\{f\}$. These feature pixels' spatial positions are then determined through epipolar geometry calculations  $\mathrm{Epip}(\cdot)$. In the actual computation process, we found that using epipolar geometry to determine depths of distance points (very low optical flow value) is highly unreliable because the two rays are nearly parallel, resulting in a calculated infinity depth. To address this, for pixels with optical flow values less than $f_{th}$, we estimate the depth of distant feature points through an approximate calculation method.  In summary, pixel $u$'s depth $d_u$ calculation is shown in~\cref{depth_cal}. 
$\theta_u$ represents the angle between the ray of $u$ and the camera's optical axis. The derivation process of~\cref{depth_cal} can be found in the supplementary material.

\begin{align}
\label{depth_cal}
d_u = \left\{\begin{matrix}
 \mathrm{Epip}(T_t, T_{t+1},f_u) & f_u \ge  f_{th} \\
 d_{th} \cdot \frac{tan\theta_u / f_{u}}{tan\theta_{th}/f_{th}}  & f_u < f_{th}
\end{matrix}\right.
\end{align}

Following the calculations above, we obtained the initial positions of Gaussians $\{P_t\}$ that cover LiDAR's blind spots (such as the upper parts of trees/buildings and distant landscapes). For the other properties of Gaussians, we initialize them using the method described in SplaTAM~\cite{splatam}.

\subsubsection{Sphere Gaussian.}\label{sphere_gaussian_text}
Vanilla 3DGS often generates ambiguous floaters to render the sky, leading to a significant number of unnecessary Gaussians and non-conformance to multi-view consistency. This phenomenon indicates that the sky lacks geometric information and cannot be represented by Gaussians in the world coordinate system and there are no Gaussian-based methods to address the issue of sky reconstruction.

We discover that the color of the sky depends on the viewing direction and is unrelated to the vehicle's translation. Therefore, we model the sky using Gaussians attached to the surface of an enormous sphere $S$ with a radius of $R$, namely Sphere Gaussians $G_{sky}$. At any given moment, there exists only a rotational transformation between the coordinate system of $G_{sky}$ and that of the vehicle. The position property of $G_{sky}$ has only two degrees of freedom. Furthermore, we stipulate that the rotation property of $G_{sky}$ is parallel to the radial direction of the spherical surface $S$, and $G_{sky}$ has a fixed thickness along the radial direction.

As a result, $G_{sky}$ possesses only the following eight learnable properties $\{[x^{sky},z^{sky}],[s_x^{sky},s_z^{sky}],c^{sky},\alpha^{sky}\}$, where $c^{sky}$ represents $[r^{sky},g^{sky},b^{sky}]$. For the remaining non-learnable properties, we calculate $y^{sky}$ and $q^{sky}$ using \cref{sphere_gaussian}.
\begin{equation}
\begin{aligned}
\label{sphere_gaussian}
y^{sky} &= \sqrt{R^2- (x^{sky})^2 - (y^{sky})^2} , \theta = arccos(\frac{y^{sky}}{R})\\
q^{sky} &= [cos(\frac{ \theta}{2}), sin(\frac{\theta}{2})\frac{x^{sky}}{R},  sin(\frac{\theta}{2})\frac{y^{sky}}{R}, sin(\frac{\theta}{2})\frac{z^{sky}}{R}]
\end{aligned}
\end{equation}

When a new RGB frame $I_t$ arrives, we transform the ray direction vectors of $I_t$'s sky pixels from the vehicle's coordinate system to the $G_{sky}$'s coordinate system, and add new Gaussians to $G_{sky}$ at the intersection position of the ray direction and the spherical surface $S$. During the rendering process, we first convert Sphere Gaussians from $G_{sky}$'s coordinate to the vehicle's coordinate, and then utilize $T_t$ to transform these Gaussians to the world coordinate system. Subsequently, we feed them to our Hybrid RGBD Rasterizer (\cref{hybrid_rgbd_rasterizer}). This approach ensures that the Sphere Gaussians are consistent across multiple views.

\subsubsection{2D Gaussian Plane.} 
We found that in Vanilla 3DGS, Gaussians on the road surface tend to stack in multiple layers with significant overlap between each other. However, both the geometry and texture of the road surface are relatively simple, leading to a substantial redundancy of the Gaussians. Meanwhile, Vanilla 3DGS sorts Gaussians of the road surface and those of roadside landscapes together. And it's worth emphasizing that Gaussian sorting consumes up to 30\% of the total rendering process~\cite{stopthepop}. Since the road surface is always below all roadside landscapes, the Gaussians of the road surface can be completely excluded from the sorting calculation, which has great potential to accelerate rendering.

Therefore, we model the road using flattened Gaussians on the plane, i.e. 2D Gaussian Plane $G_{inlier}$. The position property of $G_{inlier}$ only has two degrees of freedom due to being confined to a plane. We define $G_{inlier}$'s rotation vector $q^{inlier}$ is parallel to the plane's normal vector and $G_{inlier}$ has a fixed thickness $s_y^{inlier}$ along $q^{inlier}$.  Thus, $G_{inlier}$ possesses the following eight learnable properties: $\{[x^{inlier},z^{inlier}],[s_x^{inlier},s_z^{inlier}],c^{inlier},\alpha^{inlier}\}$.
For the remaining non-learnable properties of $G_{inlier}$, we calculate $y^{inlier}$ and $q^{inlier}$ using \cref{plane_gaussian}.

\begin{equation}
\begin{aligned}
\label{plane_gaussian}
y^{inlier} &= \frac{-Ax-Cz-D}{B}, \theta = arccos(\frac{B}{\sqrt{A^2+B^2+C^2}}) \\
q^{inlier} &= [cos(\frac{\theta}{2}), sin(\frac{\theta}{2})cos(\theta)\frac{A}{B}, 
     sin(\frac{\theta}{2})cos(\theta), sin(\frac{\theta}{2})cos(\theta)\frac{C}{B}]
\end{aligned}
\end{equation}

Initially, we get the first RGB-LiDAR frame $\{I_{0}, L_0\}$. We calculate the feature points $P_0$ of $I_0$ using \cref{depth_cal} and get $\hat{L_0}$ by concatenating $P_0$ and $L_0$, i.e. $\hat{L_0}=\mathrm{concat}(L_0,P_0)$ . Then, we implement the plane segmentation algorithm (RANSAC~\cite{open3d}) to divide $\hat{L_0}$ into inlier $\hat{L_o}^{inlier}$ and outlier $\hat{L_o}^{outlier}$, and obtain the plane equation $Ax+By+Cz+D=0$. When a subsequent RGB-LiDAR frame $\{I_t,L_t\}$ arrives, we use the aforementioned method to obtain $\hat{L_t}$ and calculate their distances to the plane mentioned above. We set a distance threshold $D_{th}$ and classify the part of $\hat{L_t}$ as ${\hat{L_t}}^{inlier}$, whose distances to the plane are less than $D_{th}$. Then we add ${\hat{L_t}}^{inlier}$ to $G_{inlier}$. Specifically, for complex urban scenes, we select some keyframes and construct plane equations between each pair of adjacent keyframes, achieving the reconstruction of undulating road surfaces.

Our approach reduces the number of Gaussians on the road and improves the rendering speed. Additionally, by distinguishing between inliers and outliers, Gaussians on the road are no longer involved in the sorting algorithm, which will be elaborated in \cref{hybrid_rgbd_rasterizer}. 


\subsubsection{3D Gaussian.} 
We use 3D Gaussian $G_{outlier}$ to model roadside landscapes. The geometric and texture of roadside landscapes are intricate. Resembling from Vanilla 3DGS, we utilize ellipsoids-shaped Gaussians to characterize $G_{outlier}$, which ensures that the Gaussians have sufficient degrees of freedom. $G_{outlier}$ possesses 14 learnable properties $\{\mu^{outlier},s^{outlier},c^{outlier},q^{outlier},\alpha^{outlier}\}$, where position $\mu$ means $[x,y,z]$ and scale $s$ means $[s_x, s_y,s_z]$.
\subsection{Rasterization and Optimization}\label{rasterization_and_optimization_text}

\subsubsection{Hybrid RGBD Rasterizer.}\label{hybrid_rgbd_rasterizer}

We design a Hybrid RGBD Rasterizer tailored for Hybrid Gaussian Representation. We calculate the weight $f(p)$ of any point $p \in \mathbb{R}^{3}$ in 3D space according to the standard Gaussian equation \cref{weight_equation}.

\begin{equation}
\begin{aligned}
\label{weight_equation}
f(p) &= \textrm{sigmoid}(\alpha) exp(-\frac{1}{2}(p-\mu)^T(p-\mu))
\end{aligned}
\end{equation}
where $\mu$ and $\alpha$ represent the center and the opacity of Gaussians, respectively.

Our Hybrid Gaussian Rasterization contains three steps.
Firstly, we transform $G_{sky}$ to the world coordinate as described in~\cref{sphere_gaussian_text}. And for a pixel belonging to a tile, we independently evaluate the three types of Gaussians within the tile. Then, we perform sorting within $G_{sky}$ and $G_{outlier}$ individually and concatenate them together. The order from front to back is ${{G}_{outlier},{G}_{inlier},{G}_{sky}}$. Finally, we render one pixel's RGB value $C$ and depth value $D$ by directly computing the weighted sum of $N$ sorted Gaussians overlapping with this pixel, as described in \cref{weight_sum_cd}.
\begin{equation}
\begin{aligned}
\label{weight_sum_cd}
C = \sum_{i=1}^{N}c_if_i\prod_{j=1}^{i-1}(1-f_j) \\ 
D = \sum_{i=1}^{N}d_if_i\prod_{j=1}^{i-1}(1-f_j)
\end{aligned}
\end{equation}
where $c$, $d$ are the color and depth of each Gaussian and $f$ is given by \cref{weight_equation}. Our approach saved 20\% of the time by sorting three types of Gaussians separately compared to sorting them together.

\subsubsection{Optimize on Keyframe List.}

To prevent degradation of the reconstruction quality of historical frames during the online mapping process~\cite{imap}, we maintain a global keyframe list in the same spirit of~\cite{niceslam}. In each iteration, a frame is randomly selected from the keyframe list for optimization. The keyframe list consists of $K$ frames. We choose $K-2$ frames randomly from all frames that overlap with current frame based on point cloud projection, and add the current and previous frame to the list. We update the keyframe list every $n$ frames.

Then we provide details on the optimization of $\{G_{sky},G_{inlier},G_{outlier}\}$. Our loss function $\mathcal{L}$ consists of three components: photometric loss $\mathcal{L}_{RGB}$, geometric loss $\mathcal{L}_{LiDAR}$, and regularization loss $\mathcal{L}_{reg}$. 

The RGB loss $\mathcal{L}_{RGB}$ consists of $\mathcal{L}_1$ and a D-SSIM term:
\begin{align}
\mathcal{L}_{RGB} &= (1-\lambda)\mathcal{L}_1 + \lambda \mathcal{L}_{\textrm{D-SSIM}}
\end{align}

The LiDAR loss $\mathcal{L}_{LiDAR}$ is the $\mathcal{L}_1$ loss between Sparse-D (LiDAR) and the predicted depth. The regularization loss $\mathcal{L}_{reg}$ aims to improve the quality of rendered depth, including depth smoothness loss $\mathcal{L}_{smooth}$ and isotropic~\cite{gaussian-splatting-slam} loss $\mathcal{L}_{iso}$:
\begin{equation}
\begin{aligned}
\label{weight_sum}
\mathcal{L}_{smooth}&=\left|\partial_{x} d_{t}^{*}\right| e^{-\left|\partial_{x} I_{t}\right|}+\left|\partial_{y} d_{t}^{*}\right| e^{-\left|\partial_{y} I_{t}\right|} \\
\mathcal{L}_{iso} &= (s-\bar{s})^T(s-\bar{s})\\
\mathcal{L}_{reg} &= \lambda_{smooth}\mathcal{L}_{smooth} + \lambda_{iso}\mathcal{L}_{iso}
\end{aligned}
\end{equation}

We optimize Gaussians by minimizing the loss $\mathcal{L}$:
\begin{align}
\underset{ \{ G_{sky},G_{inlier},G_{outlier}  \}  }{\textrm{min}} \lambda_{RGB} \mathcal{L}_{RGB} + \lambda_{LiDAR} \mathcal{L}_{LiDAR} +  \lambda_{reg} \mathcal{L}_{reg}
\end{align}

\subsection{Adaptive Update of Gaussians}\label{adaptive_update_method_text}

\subsubsection{Silhouette Filter.}\label{silhouette_filter}
We render a silhouette image $S$ according to \cref{silhouette_render}, which can reflect the density of Gaussians at different positions on the image. The silhouette image serves two purposes: adding new Gaussians in regions where Gaussians are relatively sparse, and filtering depth to address the issue of irregular spreading of depth at object edges.
\begin{align}
\label{silhouette_render}
S = \sum_{i=1}^N f_i \prod_{j=1}^{i-1} (1-f_j)
\end{align}

When a new frame $\{I_t, L_t\}$ arrives, we add it to our Hybrid Gaussian. However, our goal is to add new Gaussians only in regions where current Gaussians are relatively sparse. Therefore, we compute the silhouette and only add Gaussians where $S < S_{th}$ or $\mathcal{L}_1(D) > 50\mathrm{MDE}$.

\begin{figure}[t]
    \centering
    \begin{subfigure}[b]{\textwidth}
        \centering
        \includegraphics[width=\textwidth]{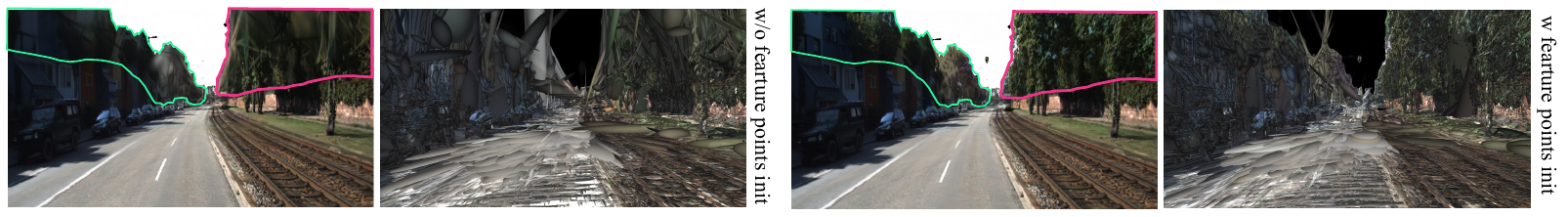}
        \caption{Effect of Gaussian Initialization.}
        \label{ablation_gaussian_init}
    \end{subfigure}
    
    
    \begin{subfigure}[b]{\textwidth}
        \centering
        \includegraphics[width=\textwidth]{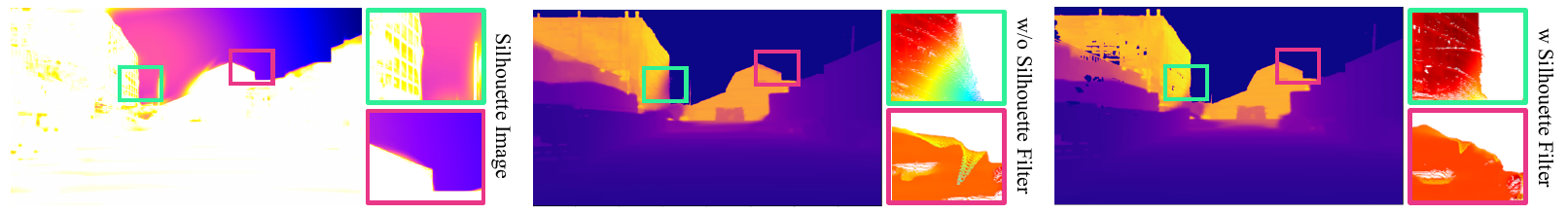}
        \caption{Effect of Silhouette Filter.}
        \label{ablation_silhouette_filter}
    \end{subfigure}
    \caption{\textbf{Effect of Gaussian Initialization \& Silhouette Filter.} 
    \cref{ablation_gaussian_init} shows that Gaussian Initialization enhances the reconstruction quality in the upper parts of trees and buildings. 
    \cref{ablation_silhouette_filter} shows that Silhouette Filter removes the abnormal depth of building's edges.}
    \label{fig_3_all}
    \vspace{-0.6cm}
\end{figure}

Moreover, we expand the application of the silhouette by using it to filter the depth map. When rendering the depth map, we observed the issue of depth shrinkage towards the optical center of the camera, as shown in \cref{ablation_silhouette_filter}. This occurrence is attributed to \cref{weight_equation}, where Gaussians are relatively sparse at object edges, and opacity decays with distance from the Gaussian center, leading to the phenomenon of depth shrinkage at object edges.


\subsubsection{Densify Control.}

Gaussian densification accomplishes the purpose of cloning new Gaussians in regions lacking geometric features and splitting large Gaussians into smaller Gaussians. This method refines the scene details by densifying Gaussians with large gradients, resolving the under-reconstruction problem~\cite{3DGS} of Gaussians. Although it enhances the quality of reconstruction, it also leads to an increase in the number of Gaussians. Therefore, for every 20 iterations, we execute densification and prune Gaussians with $\alpha$ values lower than $\alpha_{th}$ or Gaussians with $s$ larger than $s_{th}$.

\subsubsection{Importance Pruning.}
Solely utilizing opacity and scale as criteria for pruning Gaussians is inefficient. The reason is that it only ensures the majority of Gaussians eventually meet the threshold ranges we set for opacity and scale properties, but it does not guarantee the Gaussians within the threshold ranges can effectively represent the entire scene. 
To this end, we compute the importance score $IS_j$ of Gaussian $G_j$ and periodically prune Gaussians whose importance score lies in the bottom $\eta \%$. We incorporate the gradient of Gaussians $grad(\cdot)$ into the importance calculation. Gaussians with larger gradients indicate poorer reconstruction at their respective positions, thus importance is positively correlated with the gradient. We normalize the gradient term using the loss of the current frame $\mathcal{L}$. The calculation of the importance score is illustrated in \cref{Gaussian_significance}.

\begin{equation}
\begin{aligned}
\label{Gaussian_significance}
IS_j &= \sum_{i=1}^{KHW} \mathbb{I}(G_j,r_i)\cdot \alpha_j \cdot \tau_j \cdot \frac{grad_j}{\mathcal{L}}\\
\tau_j &= \mathrm{max}(\mathrm{min}(V_j,0)),V_{\mathrm{max} 50})
\end{aligned}
\end{equation}
Where $j$ and $K$ represent the index of the Gaussian and the length of keyframe list respectively. $\mathbb{I}$ indicates whether rays $r_i$ emitted from an image of size $H \times W$ intersect with Gaussian $G_j$. Besides, $V_j$ means the volume of $G_j$ and $V_{\mathrm{max50}}$ represents the 50\% largest volume of all sorted Gaussians.

\section{Experiments}
\subsubsection{Datasets \& Metrics.}
To evaluate the reconstruction quality under different lighting, weather, and road enviroments, we select three most commonly used datasets for urban scenes (KITTI~\cite{kitti}, nuScenes~\cite{nuscenes}, Waymo~\cite{waymo}) as well as one simulated dataset (VKITTI2~\cite{vkitti2}). We use their RGB images, LiDAR points, and pose data as inputs of our experiments and we select multiple challenging scenarios. To quantitatively validate the effect of our method, we employ three widely recognized photometric rendering quality metrics (PSNR, SSIM, LPIPS).

\subsubsection{Compared Baselines.}
We compared our method with two NeRF-based methods (Mip-NeRF360~\cite{mipnerf360}, Instant-NGP~\cite{instantngp}) and two Gaussian-based methods (3DGS~\cite{3DGS}, SplaTAM~\cite{splatam}). Among them, Mip-NeRF 360, InstantNGP, and 3DGS are recognized for their high-quality reconstruction results, but their training process is off-line and time-consuming. SplaTAM, currently at the forefront of Gaussian-based methods, supports online training and achieves outstanding reconstruction results indoors. To address the inability to represent the sky in street scenes, we used Deeplabv3~\cite{deeplabv3} to mask the sky before calculating metrics. Additionally, we utilize LiDAR points to initialize Gaussians, ensuring consistent initial conditions.

\subsection{Comparisons and Analysis}

\subsubsection{RGB Results.}

To validate the rendering RGB quality, we compare our method with two SOTA NeRF-based methods and two cutting-edge Gaussian-based methods. For Gaussian-based methods, 3DGS, SplaTAM, and Ours are all initialized based on sparse LiDAR points.

\begin{figure}[tbp] 
    \vspace{-0.2cm}
    \centering 
    \includegraphics[width=\textwidth]{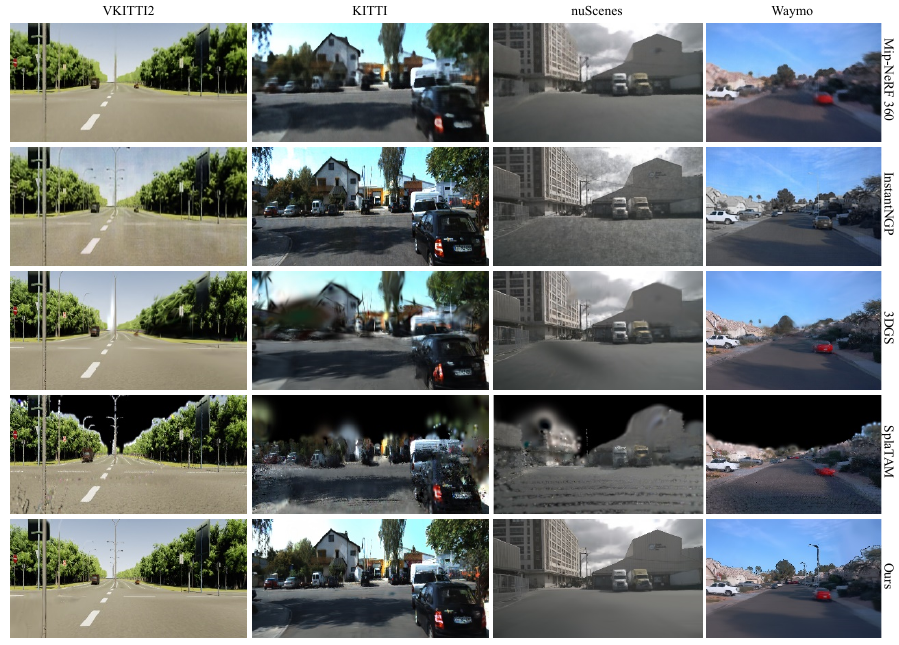} 
    \vspace{-0.4cm}
    \caption{\textbf{Quantitative RGB results in diverse urban scenes.} We compare HGS-Mapping with three off-line reconstruction methods~\cite{3DGS,instantngp,mipnerf360} and an online dense mapping method~\cite{splatam}. The three offline methods are trained~\textbf{over 20K iterations} but online methods only train 100 iterations per-frame. The results shows that our work achieves superior high-quality rendering while faster training speed.}
    \label{comparasion_RGB} 
    \vspace{-0.4cm}
\end{figure}

For offline methods, Mip-NeRF 360 is trained for several hours on the entire dataset, while InstantNGP is trained for 15 minutes (35K iterations), and 3DGS is trained for 10 minutes (20K iterations). For online methods (SplaTAM, Ours), we train 100 iterations per frame. All results are reported on a single RTX3090.

\cref{AllRenderTable} shows that even when compared with offline methods, our rendering quality is significantly better. In urban scenes, our method outperforms SplaTAM by a large margin in terms of rendering quality. \cref{comparasion_RGB} illustrates SplaTAM's lack of geometric reconstruction ability in areas without initialized LiDAR points (upper part of trees/buildings and sky).

The rendering quality of SplaTAM in LiDAR coverage areas is also unsatisfactory. Although SplaTAM performs well in indoor scenes with dense depth using a large number of small-sized spheres, its ability to fill in unreconstructed areas with sparse LiDAR points initialization outdoors is inferior to that of ellipsoids. This will be further elaborated in the supplementary material.

\begin{table}[t]
\centering
\caption{\textbf{Quantitative results of RGB rendering} on four urban datasets \cite{vkitti2,kitti,nuscenes,waymo}.}
\vspace{-0.3cm}
\label{AllRenderTable}
\begin{tabular}{c|cc|cc|cc|cc} 
\hline
                & \multicolumn{2}{c|}{VKITTI2}                                                  & \multicolumn{2}{c|}{KITTI}                                                    & \multicolumn{2}{c|}{nuScenes}                                                 & \multicolumn{2}{c}{Waymo}                                                      \\
                & PSNR$\uparrow$                                  & SSIM$\uparrow$                                  & PSNR$\uparrow$                                  & SSIM$\uparrow$                                  & PSNR$\uparrow$                                  & SSIM$\uparrow$                                  & PSNR$\uparrow$                                  & SSIM$\uparrow$                                   \\ 
\hline
Mip NeRF 360    & {\cellcolor[rgb]{1,0.8,0.529}}24.878  & {\cellcolor[rgb]{1,0.98,0.529}}0.7502 & {\cellcolor[rgb]{1,0.98,0.529}}21.395 & {\cellcolor[rgb]{1,0.98,0.529}}0.6499 & {\cellcolor[rgb]{1,0.8,0.529}}28.530  & {\cellcolor[rgb]{1,0.98,0.529}}0.8751 & {\cellcolor[rgb]{1,0.98,0.529}}24.826 & 0.8440                                 \\
Instant-NGP     & 21.586                                & 0.6305                                & {\cellcolor[rgb]{1,0.6,0.6}}22.530    & {\cellcolor[rgb]{1,0.8,0.529}}0.7418  & 27.046                                & 0.8352                                & {\cellcolor[rgb]{1,0.8,0.529}}26.154  & {\cellcolor[rgb]{1,0.8,0.529}}0.8557   \\
3DGS(LIDAR)     & {\cellcolor[rgb]{1,0.98,0.529}}24.236 & {\cellcolor[rgb]{1,0.8,0.529}}0.7928  & 19.227                                & 0.6185                                & {\cellcolor[rgb]{1,0.98,0.529}}28.091 & {\cellcolor[rgb]{1,0.8,0.529}}0.9133  & 17.664                                & {\cellcolor[rgb]{1,0.98,0.529}}0.8491  \\
Splatam(masked) & 20.003                                & 0.7401                                & 14.778                                & 0.4783                                & 18.998                                & 0.7459                                & 20.371                                & 0.7704                                 \\
Ours            & {\cellcolor[rgb]{1,0.6,0.6}}29.114    & {\cellcolor[rgb]{1,0.6,0.6}}0.9019    & {\cellcolor[rgb]{1,0.8,0.529}}22.520  & {\cellcolor[rgb]{1,0.6,0.6}}0.8007    & {\cellcolor[rgb]{1,0.6,0.6}}30.862    & {\cellcolor[rgb]{1,0.6,0.6}}0.9404    & {\cellcolor[rgb]{1,0.6,0.6}}26.445    & {\cellcolor[rgb]{1,0.6,0.6}}0.8832     \\
\hline
\end{tabular}
\vspace{-0.2cm}
\end{table}

\subsubsection{Depth Results.}

\begin{figure}[htbp] 
    \centering 
    \vspace{-0.4cm}
    \includegraphics[width=\textwidth]{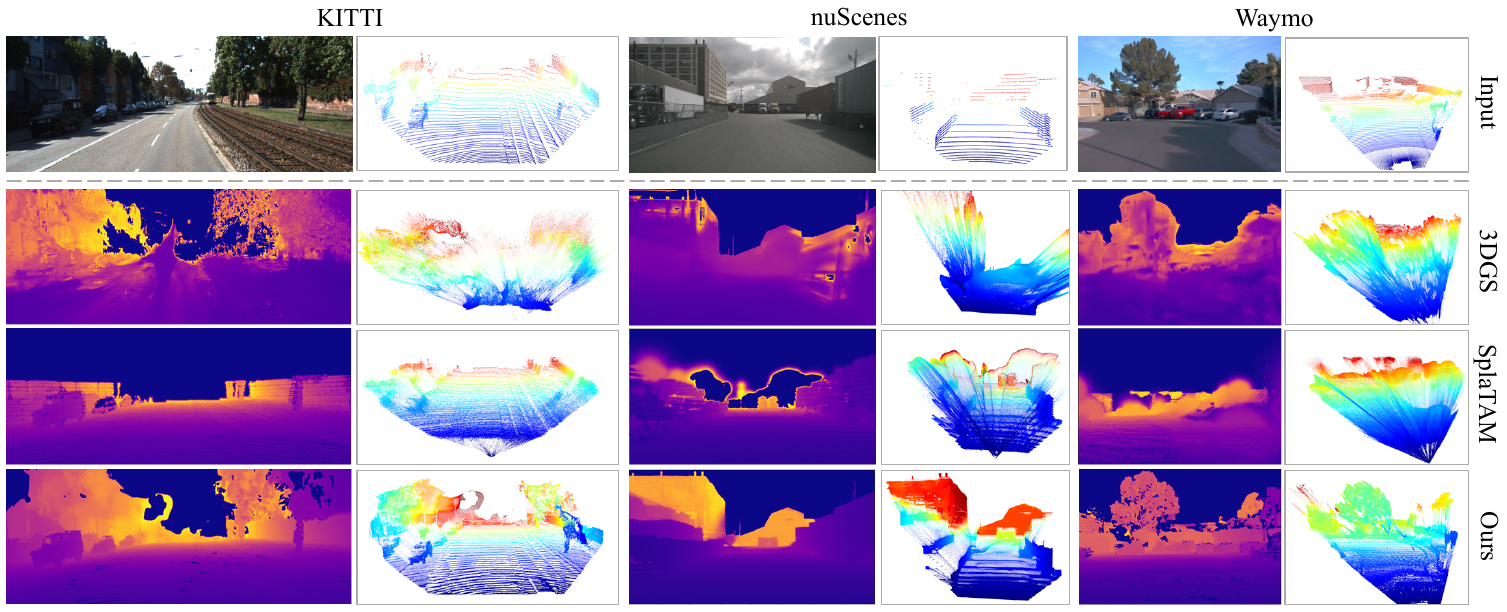} 
    \caption{\textbf{Qualitative depth results} on KITTI~\cite{kitti}, nuScenes~\cite{nuscenes} and Waymo~\cite{waymo}.} 
    \label{comparason_depth} 
    \vspace{-0.9cm}
\end{figure}

We compare the depth images of 3DGS, SplaTAM, and ours. Following the settings in the previous section, we render the depth and projected dense depth into point clouds for better visualization. Vanilla 3DGS lacks the capability to render depth, we load 3DGS's checkpoint into SplaTAM's rasterizer to enable depth rendering. Due to the absence of ground truth dense depth in urban scenes, inspired by the depth completion task~\cite{depthcompletion}, we record the Mean Absolute Error (MAE) and Root Mean Square Error (RMSE) between the predicted depth and sparse LiDAR data to assess the quality of depth map, that is shown in \cref{comparison_depth_table}. It is important to note that these two metrics can only evaluate the quality of the depth within the LiDAR coverage. \cref{comparason_depth} demonstrates that our method can render high-quality dense depth maps relying on $\mathcal{L}_{LiDAR}$ and $\mathcal{L}_{smooth}$. We are surprised to find that even for the very sparse LiDAR points in nuScenes dataset (only $\frac{2975}{1600\times 900}$ pixels have valid depth), our method can clearly discern the geometric shapes of distant buildings and vehicles.

\begin{table}
\centering
\caption{\textbf{Sparse depth comparison} using depth completion metrics\cite{depthcompletion}.}
\vspace{-0.4cm}
\label{comparison_depth_table}
\begin{tblr}{
  cells = {c},
  cell{1}{2} = {c=2}{},
  cell{1}{4} = {c=2}{},
  cell{1}{6} = {c=2}{},
  vline{2-3,5} = {1}{},
  vline{2,4,6} = {1-4}{},
  hline{1,3,5} = {-}{},
}
        & KITTI           &                  & nuScenes        &                  & Waymo           &                  \\
        & MAE$\downarrow$ & RMSE$\downarrow$ & MAE$\downarrow$ & RMSE$\downarrow$ & MAE$\downarrow$ & RMSE$\downarrow$ \\
SplaTAM & 0.503           & 0.973            & 0.079           & 1.563            & 0.353           & 1.819            \\
Ours    & \textbf{0.203}         & \textbf{0.522}          & \textbf{0.033}           & \textbf{0.309 }           & \textbf{0.101  }         & \textbf{0.673}            
\end{tblr}
\vspace{-0.7cm}
\end{table}

\subsubsection{Mesh Results.}

\begin{figure}[htbp] 
    \vspace{-0.4cm}
    \centering 
    \includegraphics[width=\textwidth]{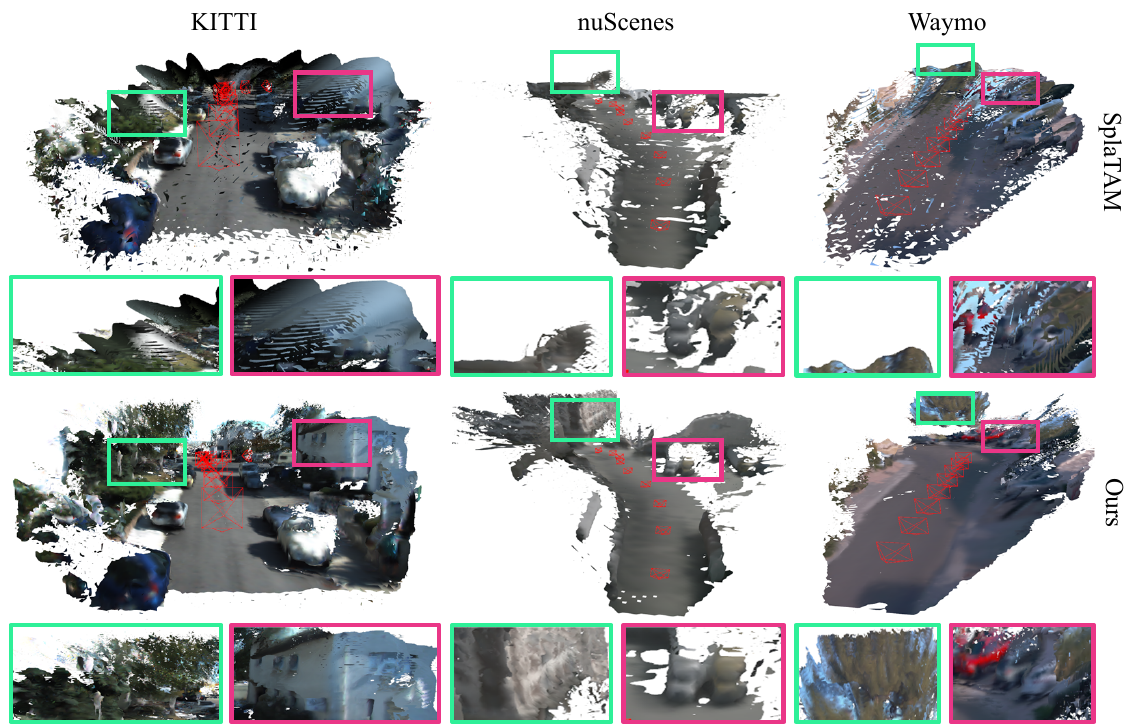} 
    \caption{\textbf{Qualitative Mesh Comparisons} on KITTI~\cite{kitti}, nuScenes~\cite{nuscenes} and Waymo~\cite{waymo}.} 
    \label{mesh_compare} 
    \vspace{-0.7cm}
\end{figure}

To provide a more intuitive comparison of the geometric reconstruction results across the entire scene, we qualitatively compare meshes of Ours and SplaTAM, as shown in \cref{mesh_compare}. We employ voxel-fusion~\cite{tsdf-fusion} and marching cubes~\cite{marchingcubes} to export the rendered dense depth map sequence as meshes. We observe that due to the limited coverage of the LiDAR, SplaTAM render very poor geometric shapes outside the LiDAR's coverage. However, owing to our feature point initialization, we can accurately reconstruct the upper parts of trees/buildings. Furthermore, due to our separate representation of the road surface and roadside objects, our road surface is smoother and devoid of holes.

\subsection{Ablation Studies}

\subsubsection{Hybrid Gaussian Representation.}

Our 3D Gaussian $G_{outlier}$'s properties are the same as Vanilla 3DGS.
Thus, we use $G_{outlier}$ to simultaneously model the sky, road surface, and roadside landscapes as a baseline. Subsequently, we progressively substitute $G_{sky}$ and $G_{inlier}$ to model the sky and road surface, respectively. To validate the effects of our Hybrid Gaussian Representation on improving rendering quality and speed, we compare five metrics, including Rendering FPS, SIZE, PSNR, etc., under the same KITTI scenario.

\begin{figure}[htbp]
    \begin{minipage}[h]{0.48\linewidth}
        \centering
        \begin{tblr}{
          cells = {c},
          vline{2} = {-}{},
          hline{1-2,6} = {-}{},
          colsep=4pt, 
        }
        $\eta$ \%                & SIZE$\downarrow$ & PSNR$\uparrow$ & SSIM$\uparrow$ & LPIPS$\downarrow$ \\
        -   &   245.4MB &   22.542   &    0.7596  &     0.189  \\
        1\% &  234.5MB    &  22.364    &   0.7215 &   0.204  \\
        3\% &   220.8MB   &   22.054  &   0.6955 &    0.230  \\
        5\% &    183.9MB  &   21.365 &  0.6479    &  0.275     
        \end{tblr}
        \captionof{table}{\textbf{Ablation on Importance Pruning} with different pruning rate $\eta\%$. As $\eta\%$ increases, storage size significantly declines, while there is a corresponding drop in rendering quality.}
        \label{importance_prune}
    \end{minipage}
    \hfill
    \begin{minipage}[h]{0.45\linewidth}
        \centering
        \includegraphics[width=\linewidth]{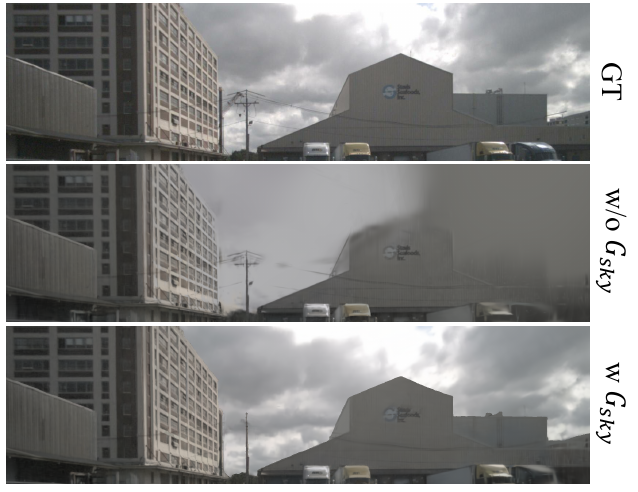}
        \captionof{figure}{Ablation on $G_{sky}$.}
        \label{ablation_sky}
    \end{minipage}
    \vspace{-0.5cm}
\end{figure}

The second row of \cref{ablation_HGS} indicates that modeling the sky using $G_{sky}$ not only reduces storage consumption but also enhances rendering speed. Furthermore, we select scenes with complex sky colors from nuScenes and conducted ablation experiments visualizing our Sphere Gaussian. \cref{ablation_sky} illustrates our $G_{sky}$ rendering high-quality sky and clouds. The last row of \cref{ablation_HGS} demonstrates that using $G_{inlier}$ results in a slight decrease in rendering quality (from 22.660 to 22.542) but achieves remarkable improvements in storage efficiency (from 298.4MB to 245.4MB) and rendering speed (from 228FPS to 271 FPS).
\begin{table}
\centering
\caption{\textbf{Ablations on Hybrid Gaussian Representation.}}
\vspace{-0.3cm}
\label{ablation_HGS}
\begin{tabular}{>{\centering\arraybackslash}p{1cm}>{\centering\arraybackslash}p{1cm}>{\centering\arraybackslash}p{1cm}>{\centering\arraybackslash}p{1.5cm}>{\centering\arraybackslash}p{1.5cm}>{\centering\arraybackslash}p{1.5cm}>{\centering\arraybackslash}p{1.5cm}>{\centering\arraybackslash}p{1.5cm}}   
\hline
$G_{outlier}$ & $G_{sky}$ & \multicolumn{1}{c}{$G_{inlier}$} & FPS$\uparrow$                   & SIZE$\downarrow$                    & PSNR$\uparrow$                     & SSIM$\uparrow$                     & LPIPS$\downarrow$                  \\ 
\hline
\ding{51}          & \ding{55}      & \ding{55}                             & 209                             & 318.9MB                             & 21.246                             & 0.6175                             & 0.296                              \\
\ding{51}          & \ding{51}      & \ding{55}                             & {\cellcolor[rgb]{1,0.8,0.6}}228 & {\cellcolor[rgb]{1,0.8,0.6}}298.4MB & {\cellcolor[rgb]{1,0.6,0.6}}22.660 & {\cellcolor[rgb]{1,0.6,0.6}}0.7621 & {\cellcolor[rgb]{1,0.6,0.6}}0.182  \\
\ding{51}          & \ding{51}      & \ding{51}                             & {\cellcolor[rgb]{1,0.6,0.6}}271 & {\cellcolor[rgb]{1,0.6,0.6}}245.4MB & {\cellcolor[rgb]{1,0.8,0.6}}22.542 & {\cellcolor[rgb]{1,0.8,0.6}}0.7596 & {\cellcolor[rgb]{1,0.8,0.6}}0.189  \\
\hline
\end{tabular}
\vspace{-0.3cm}
\end{table}

\subsubsection{Importance Pruning.}
When the length of the keyframe list equals $K$, we begin to record global significance every 5 iterations, and perform global significance pruning on $G_{outlier}$ every 2 frames, removing $\eta\%$ of the $G_{outlier}$ each time. We conduct experiments with different importance pruning rates $\eta\%$, namely turning it off and setting it to 1\%, 3\%, 5\%, as shown in \cref{importance_prune}. Increasing the pruning rate effectively reduces storage size, but setting it too high can lead to a significant decrease in rendering quality.

\subsubsection{Gaussian Initialization \& Silhouette Filter.}

Regarding Feature Points Initialization in \cref{gaussian_init}, we conduct tests in scenes with abundant trees and extensive visibility. The experimental results, as depicted in \cref{ablation_gaussian_init}, indicate that our initialization method significantly enhances the reconstruction performance in the upper parts of trees and buildings.

As for the Silhouette Filter in \cref{silhouette_filter}, \cref{ablation_silhouette_filter} shows that the silhouette filter effectively resolves the issue of object edge recession in the depth map.

\section{Conclusion}
We are the first 3DGS-based online dense mapping framework in urban scenes by proposing the novel Hybrid Gaussian Representation suited for complex unbounded scenes. Additionally, extensive experiments demonstrate that our representation and optimization methods significantly improve rendering speed and quality, achieving SOTA performance. However, the efficacy of the RANSAC method for extracting $G_{inlier}$ becomes limited under conditions such as rugged roads or significant curvature. Therefore, there remains potential for further enhancement of this framework to accommodate arbitrary outdoor scenes.

%
%
\newpage
\bibliographystyle{splncs04}
\bibliography{egbib}
\end{document}